\pdfoutput=1
\documentclass{article}
\usepackage{graphicx}
\usepackage{hyperref}
\hypersetup{colorlinks,allcolors=black}
\usepackage[final]{corl_2019} % Uncomment for the camera-ready ``final'' version
\title{Hindsight Experience Replay with Kronecker Product Approximate Curvature}

\author{
  Dhuruva Priyan G M\\
  Amrita School of Engineering, Coimbatore\\
  Amrita Vishwa Vidhyapeetham, India\\ 
  \texttt{gmdhuruva@gmail.com} \\
   \And
   Abhik Singla \\
   Dept of Computer Science and Automation \\
   Indian Institute of Science, Bangalore \\
   \texttt{abhiksingla@iisc.ac.in} \\
   \AND
   Shalabh Bhatnagar \\
   Dept of Computer Science and Automation \\
   Indian Institute of Science, Bangalore \\
   \texttt{shalabh@iisc.ac.in} \\
}
\begin{document}
\maketitle

%===============================================================================

\begin{abstract}
Hindsight Experience Replay (HER) is one of the efficient algorithm to solve Reinforcement Learning tasks related to sparse rewarded environments. But due to its reduced sample efficiency and slower convergence HER fails to perform effectively. Natural gradients solves these challenges by converging the model parameters better. It avoids taking bad actions that collapse the training performance. However updating parameters in neural networks requires expensive computation and thus increase in training time. Our proposed method solves the above mentioned challenges with better sample efficiency and faster convergence with increased success rate.   A common failure mode for DDPG is that the learned Q-function begins to dramatically overestimate Q-values, which then leads to the policy breaking, because it exploits the errors in the Q-function. We solve this issue by including Twin Delayed Deep Deterministic Policy Gradients (TD3) in HER. TD3 learns two Q-functions instead of one and it adds noise to the target action, to make it harder for the policy to exploit Q-function errors. The experiments are done with the help of OpenAi’s Mujoco environments. Results on these environments show that our algorithm (TDHER+KFAC) performs better in most of the scenarios.
\end{abstract}


%===============================================================================

\section{Introduction}
	
Multi goal sparse rewarded environments is one of the challenges in Deep Reinforcement Learning. Hindsight Experience Replay tries to solves this  problem by introducing synthetic data to the replay buffer. Thus inproving overall sample efficiency of the model. HER method makes use of deep neural networks to represent control policies. Despite the impressive results, these neural networks are still trained using simple variants of stochastic gradient descent (SGD). SGD and related first-order methods explore weight space inefficiently. Sample efficiency is a major concern in RL; robotic interaction with the real world is typically scarcer than computation time, and even in simulated environments the cost of simulation often dominates that of the algorithm itself. One way to effectively reduce the sample size is to use more advanced optimization techniques for gradient updates. Natural policy gradient  [11] uses the technique of natural gradient descent to perform gradient updates. Natural gradient methods follow the steepest descent direction that uses the Fisher metric as the underlying metric, a metric that is based not on the choice of coordinates but rather on the manifold (i.e., the surface).Kronecker-factored approximated curvature (K-FAC) [1] is a scalable approximation to natural gradient. It has been shown to speed up training of various state-of-the-art large-scale neural networks [2] in supervised learning by using larger mini-batches. Unlike HER, each update is comparable in cost to an SGD update, and it keeps a running average of curvature information, allowing it to use small batches. This suggests that applying K-FAC to policy optimization could improve the sample efficiency of the current deep RL methods. In this paper, we introduce the Twin Delayed HER itk Kronecker Factored approximation Curvature method, a scalable DDPG optimization algorithm for actor-critic methods. The proposed algorithm uses a Kronecker-factored approximation to natural policy gradient that allows the covariance matrix of the gradient to be inverted efficiently. To best of our knowledge, we are also the first to extend the natural policy gradient algorithm to optimize value functions via Gauss-Newton approximation. In practice, the per-update computation cost of HER is only 10\% to 25\% higher than SGD-based methods. Empirically, we show that HER substantially improves both sample efficiency and the final performance of the agent in the MuJoCo [27] tasks compared to the vannila HER.

\section{Background}
\label{sec:background}
Multi goal sparse rewarded environments is one of the challenges in Deep Reinforcement Learning. Hindsight Experience Replay tries to solves this  problem by introducing synthetic data to the replay buffer. Thus improving overall sample efficiency of the model. HER methods make use of deep neural networks to represent control policies. Despite the impressive results, these neural networks are still trained using simple variants of stochastic gradient descent (SGD). SGD and related first-order methods explore weight space inefficiently. Sample efficiency is a major concern in RL; robotic interaction with the real world is typically scarcer than computation time, and even in simulated environments the cost of simulation often dominates that of the algorithm itself. One way to effectively reduce the sample size is to use more advanced optimization techniques for gradient updates. Natural policy gradient  [11] uses the technique of natural gradient descent to perform gradient updates. Natural gradient methods follow the steepest descent direction that uses the Fisher metric as the underlying metric, a metric that is based not on the choice of coordinates but rather on the manifold (i.e., the surface).Kronecker-factored approximated curvature (K-FAC) [1] is a scalable approximation to natural gradient. It has been shown to speed up training of various state-of-the-art large-scale neural networks [2] in supervised learning by using larger mini-batches. Unlike HER, each update is comparable in cost to an SGD update, and it keeps a running average of curvature information, allowing it to use small batches. This suggests that applying K-FAC to policy optimization could improve the sample efficiency of the current deep RL methods. In this paper, we introduce the Twin Delayed HER itk Kronecker Factored approximation Curvature method, a scalable DDPG optimization algorithm for actor-critic methods. The proposed algorithm uses a Kronecker-factored approximation to natural policy gradient that allows the covariance matrix of the gradient to be inverted efficiently. To best of our knowledge, we are also the first to extend the natural policy gradient algorithm to optimize value functions via Gauss-Newton approximation. In practice, the per-update computation cost of HER is only 10\% to 25\% higher than SGD-based methods. Empirically, we show that HER substantially improves both sample efficiency and the final performance of the agent in the MuJoCo [27] tasks compared to the vannila HER.
	\subsection{Literature Review}
	 Kronecker Factored Approximation Curvature [1] for neural networks was first done by (R. Grosse and J. Martens 2015) and they extended it for convolutional layers. Later on KFAC for recurrent Neural Networks [2] was is carried on by (James Martens, Jimmy Ba, Matt Johnson,2018). Scalable trust-region method for deep reinforcement learning using kronecker-factored approximation[3] was the first efforts to include KFAC in Reinforcement Learning algorithms. ACKTR the short form of their algorithm, gave promising results in actor-critic methods. (Jiaming Song , Yuhuai Wu 2018) analysed the results of KFAC on Proximal Policy Optimization algorithm. Hindsight Experience Replay [5] was one of the major break throughs in multi goal RL algorithms. This is a off policy algorithm which work good in sparse rewarded environments. The On policy version of this is called Hindsight Policy Gradients (2017)[6]. There were many attempt to improve the sample efficiency of HER Visual Hindsight Experience Replay (2019) [7] increase sample efficiency through re-imagining unsuccessful trajectories as successful ones by replacing the originally intended goals. Overcoming Exploration in Reinforcement Learning with Demonstrations [8] was another notable work to solve sample efficiency problem. Multi-goal reinforcement learning: Challenging robotics environments and request for research [9] provides set of concrete research ideas for improving RL algorithms, most of which are related to MultiGoal RL and Hindsight Experience Replay.
	 \subsection{Reinforcement Learning for Goal Based Environments:}
	 Reinforcement learning refers to goal-oriented algorithms, which learn how to attain a complex objective (goal) or maximize along a particular dimension over many steps; for example, maximize the points won in a game over many moves. They can start from a blank slate, and under the right conditions they achieve superhuman performance.These algorithms are penalized when they make the wrong decisions and rewarded when they make the right ones. A RL algorithm consitutes the following Keywords: Agents, Environments, States, Actions and Rewards. Agent: An agent takes actions; for example, a drone making a delivery. Action (A): A is the set of all possible moves the agent can make. Environment: The world through which the agent moves. State (S): A state is a concrete and immediate situation in which the agent finds itself. Reward (R): A reward is the feedback by which we measure the success or failure of an agent’s actions. Policy (P): The policy is the strategy that the agent employs to determine the next action based on the current state. Value (V): The expected long-term return with discount, as opposed to the short-term reward R. Q-value or action-value (Q): Q-value is similar to Value, except that it takes an extra parameter, the current action(A). Trajectory: A sequence of states and actions that influence those states. Discount factor: The discount factor is multiplied by future rewards as discovered by the agent in order to dampen these rewards’ effect on the agent’s choice of action.
	\subsection{Deep Deterministic Policy Gradients:}
	Deep Deterministic Policy Gradients (DDPG) (Lillicrap et al., 2015) is a model-free RL algorithm for continuous action spaces. Here we sketch it only informally, see Lillicrap et al. (2015) for more details. In DDPG we maintain two neural networks: a target policy (also called an actor) P : S → A and an action-value function approximator (called the critic) Q : S × A → R. The critic’s job is to approximate the actor’s action-value function QP . Episodes are generated using a behavioral policy which is a noisy version of the target policy, e.g. Pb(s) = P(s) + N (0, 1). The critic is trained in a similar way as the Q-function in DQN but the targets yt are computed using actions outputted by the actor, i.e. yt = rt + Q(st+1, P(st+1)). The actor is trained with mini batch gradient descent on the loss L = -Esq(s,P(s)), where s is sampled from the replay buffer. The gradient of L w.r.t. actor parameters can be computed by backpropagation through the combined critic and actor networks.
	\subsection{Twin Delayed DDPG:}
	TD3 is an improved version of DDPG.The three main updates to DDPG algorithm are (1) Clipped Double Q Learning (2) Delayed Policy Updates (3) Target Policy Smoothing. TD3 trains a deterministic policy in an off-policy way. Because the policy is deterministic, if the agent were to explore on-policy, in the beginning it would probably not try a wide enough variety of actions to find useful learning signals. To make TD3 policies explore better, we add noise to their actions at training time, typically uncorrelated mean-zero Gaussian noise.[12]
	\subsection{Hindsight Experience Replay}
	Hindsight Experience Replay is an off-policy RL algorithm which is used to effectively in sparse and binary rewarded environments. It uses DDPG algorithm for training Actor and Critic models.Generally, in off-policy learning algorithm an replay buffer is used to store the trajectory of our agent intracting with the environment. Inaddition to existing episodes  HER stores synthetic data to improve sample efficiency.Example consider an agent performs an episode of trying to reach goal state G from initial state S, but fails to do so and ends up in some state S’ at the end of the episode. We cache the trajectory into our replay buffer:  where r with subscript k is the reward received at step k of the episode, and a with subscript k is the action taken at step k of the episode. The idea in HER is to imagine that our goal has actually been S’ all along, and that in this alternative reality our agent has reached the goal successfully and got the positive reward for doing so. So, in addition to caching the real trajectory as seen before, we also cache the following trajectory : This synthetic trajectory helps HER to learn from failures.
	\begin{figure}
    \centering
    \includegraphics[width=10cm]{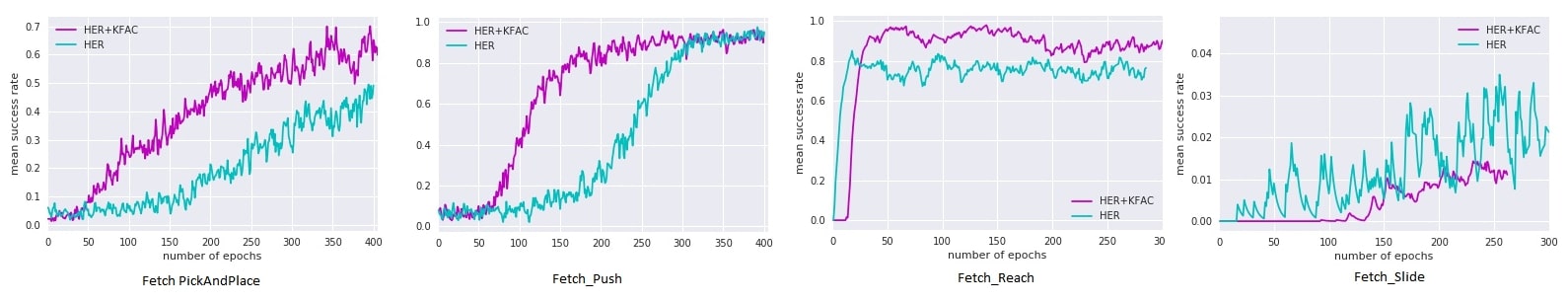}
    \caption{The algorithm is tested in OpenAI's Robotics environments. The graph is plotted against mean success rate vs number of epochs. In Most of the environments the algorithm learns within 100 epochs which clearly indicates the sample efficiency of it.}
    \label{fig:my}
\end{figure}
	\section{Method}
In conventional HER, all the above network uses Stochastic Gradient Descent (SGD) with Adam optimizer for updating the neural network parameters. Whereas Our method uses Approximated Natural Gradient Descent which acquire the loss landscape correctly by using Fisher information matrix (FIM) as a curvature of loss function and converges faster in term of ‘iterations’ than a simple first-order method (SGD).
	\subsection{HER with KFAC:}
	HER uses Deep Deterministic Policy Gradients(DDPG) algorithm. The DDPG algorithm uses four neural networks (1) Q network (2) Deterministic policy function (3) Target Q Network (4) Target policy Network. All the above network uses Stochastic Gradient Descent (SGD) with Adam optimizer for updating the neural network parameters. All of these neural network parameters is optimized with KFAC.
	\subsubsection{Natural Gradients with Deep Deterministic Policy Gradients:}
	Actor (Policy) \& Critic (Value) Network Updates The value network is updated similarly as is done in Q-learning. The updated Q value is obtained by the Bellman equation: 					However, in DDPG, the next-state Q values are calculated with the target value network and target policy network. Then, we minimize the mean-squared loss between the updated Q value and the original Q value:					For the policy function, our objective is to maximize the expected return:						The update rule for Natural gradient is 							where is the Inverse of Fisher Information Matrix. Computing is expensive. So K-FAC approximates each block as a product of two matrix and approximating them to a block diagonal.If the actor and critic are disjoint, one can separately apply K-FAC updates to each using the metrics defined above. But since in DDPG Actor networks learn from Critic network. We need to update KFAC weights simultanously. The KFAC updates are happened in the following ways:For actor and critic networks, KFAC starts computing the gradients of the weight values in the networks.The to compute and apply Kronecker Products we need to sample the loss function.The loss function is sampled by adding a random noise to it. After computing the gradients and inverting the Fisher matrix the gradients are applied back to the networks.
	\subsubsection{Twin Delayed HER + KFAC:}TD3 learns from two Q functions rather than DDPG with one Q function. And the smaller of the two Q-values to form the targets in the Bellman error loss functions. Updating the two Q networks with KFAC increases computation. The gradients are computed and applied as in the case of DDPG. TD3 updates the policy (and target networks) less frequently than the Q-function. KFAC updates the policy network once for every Q-function update.TD3 adds noise to the target action, to make it harder for the policy to exploit Q-function errors by smoothing out Q along changes in action.
	
%===============================================================================

\section{Experimental Results}
\label{sec:result}

	In this section we compare the results of all goal based algorithms. Each environment description is given along with the performance of the algorithm in those environments. The algorithms are tested in OpenAI's Robotics  environments. All the experiements trained on Intel Core™ i7-4770 CPU @ 3.40GHz × 8 cores and 1mpi cpu.
	
	\begin{figure}
    \centering
    \includegraphics[width=10cm]{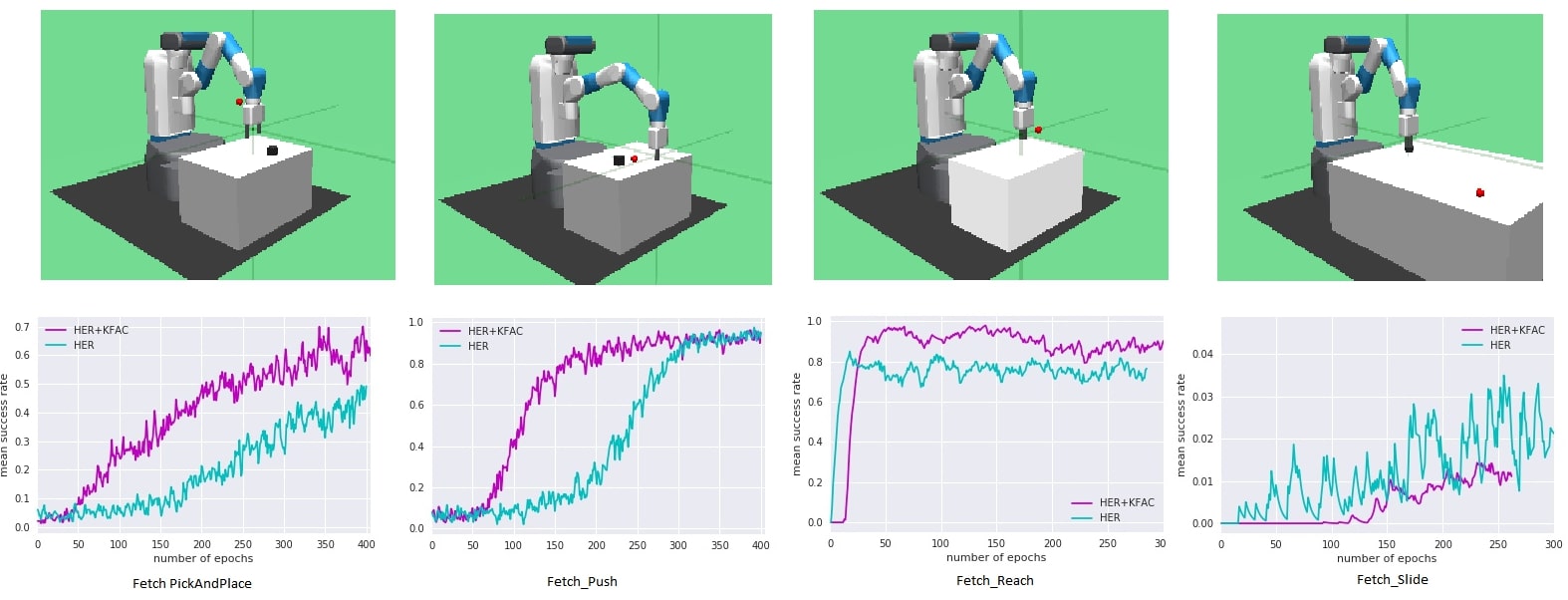}
    \caption{The Algorithm is tested in The above four OpenAI's mujoco environments:[1]Fetch PickAndPlace [2]Fetch Push [3]Fetch Reach [4]Fetch Slide.The above results shows that our algorithm outperforms the Vanilla HER.}
    \label{fig:my_labe}
\end{figure}

	\subsection{FetchPush:}A box is placed on a table in front of the robot and the task is to move it to a target location on the table. The robot fingers are locked to prevent grasping. The learned behavior is usually a mixture of pushing and rolling. The results on FetchPush environments depict that HER when combined with KFAC shows better sample efficiency. Our algorithm started learning within 50 epochs whereas vanilla HER around 200 epoch. The reason for this improvement is that natural gradient finds the optimal path of steepest descent than  SGD.
	\subsection{FetchPickAndPlace:}The task is to grasp a box and move it to the target location which may be located on the table surface or in the air above it. This environment is the perfect example of multi goal RL.This environment involves solving to goals. First to pick the box, then to place it.The above results are on of the promising results of HER+KFAC. Our algorithm reached the mean accuracy rate of 40\% of accuracy within 150 epochs whereas HER not even reached 10\%.
	\subsection{FetchReach:}The task is to move the gripper to a target position. This task is very easy to learn and is therefore a suitable benchmark to ensure that a new idea works at all. The results on FetchReach environments shows that our algorithm reaches improved accuracy.In Fetch Reach environment our algorithm shows improved accuracy over existing HER algorithm.
	\subsection{FetchSlide:}A puck is placed on a long slippery table and the target position is outside of the robot’s reach so that it has to hit the puck with such a force that it slides and then stops at the target location due to friction.The results show the training instability in this environment. HER+KFAC does't perform better in FetchSlide. The reason for this is that KFAC not able to effectively learn from the trajectories stored in the experience buffer. Fetch Slide is a multi step environment. there are two kinds of failure trajectories stored. Firstly, if the gripper fails to find the puck on the floor and Secondly,  After hitting the puck and the puck not reached the goal state. In both of these failure cases we store the hindsight, But the agent not able to clearly differentiate between the first goal and second one and learn from that. This causes the KFAC not to perform better in this environments.
		\begin{figure}
    \centering
    \includegraphics[width=10cm]{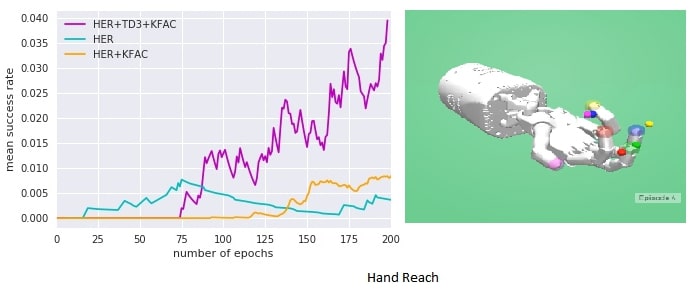}
    \caption{The algorithm is tested on Hand Environments. In this environments HER+Kfac+TD3 performs better than HER+kfac.It achieves 4\% within 200 Epochs.Our algorithm is the first to set benchmark of this accuracy rate with sample efficiency in Hand Reach environment.}
    \label{fig:my_lel}
\end{figure}
	\subsection{HandReach:}A simple task in which the goal is 15-dimensional and contains the target Cartesian position of each fingertip of the hand. A goal is considered achieved if the mean distance between fingertips and their desired position is less than 1 cm.HER when combined with twin delayed DDPG performs much better in hand environments. It reaches 4\% of the accuracy within 200 epochs.

	\begin{figure}
    \centering
    \includegraphics[width=7cm]{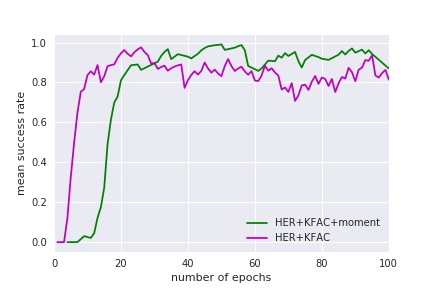}
    \caption{The algorithm is trained with several Hyperparameters. One of the hyperparameter which showed significant improvement in accuracy is plotted here. Factored Damping at the rate of 0.8 is applied on the training phase showed improved results. }
    \label{fig:my_l}
    \end{figure}
    	\subsection{HyperParameters:}Factored Tikhonov Damping:In KFAC block-diagonal approximation, Factored Damping results in adding to each of the individual diagonal blocks. Thus regularizers the parameters. The results with Fetch Reach Environment shows slightly more accuracy that HER+KFAC but takes more timesteps to learn.
	Moment:When a momentum of 0.8 is applied to KFAC, the algorithm performs slighty better then the previous version.
%===============================================================================

\section{Conclusion}
\label{sec:conclusion}

In this work we proposed a sample-efficient Hindsight Experience Replay method for deep reinforcement learning. We used a recently proposed technique called K-FAC to approximate the natural gradient update for actor-critic methods. To the best of our knowledge, we are the first to propose HER with KFAC. The algorithm had been developed and available as docker containers. The algorithms are trained with several hyperparameters and results are logged. The experiments are done with the help of OpenAi’s Mujoco environments. We trained our models in four Robotics environments namely, (1) Fetch Reach (2) Fetch Push (3) FetchPickAndPlace (4)FetchSlide. Results on these environments show that our algorithm (TDHER+KFAC) performs better in most of the scenarios. But the training time is 2.5 times the HER algorithm. Our Future work is to reduce the training time and to improve results on Fetch Slide Environments.

%===============================================================================

% The maximum paper length is 8 pages excluding references and acknowledgements, and 10 pages including references and acknowledgements

%===============================================================================


\end{document}